\documentclass{article}

\usepackage{PRIMEarxiv}

\usepackage[utf8]{inputenc} 
\usepackage[T1]{fontenc}    
\usepackage{hyperref}       
\usepackage{url}            
\usepackage{booktabs}       
\usepackage{amsfonts}       
\usepackage{nicefrac}       
\usepackage{microtype}      
\usepackage{lipsum}
\usepackage{fancyhdr}       
\usepackage{graphicx}       
\graphicspath{{media/}}     
\usepackage{caption}
\usepackage{subcaption}
\usepackage{bbm}
\usepackage{amsmath,amssymb}
\usepackage{xcolor}
\usepackage[export]{adjustbox}

\title{Local Law 144: A Critical Analysis of Regression Metrics
}

\author{
  Giulio Filippi, Sara Zannone\thanks{contact: sara.zannone@holisticai.com} , Airlie Hilliard, Adriano Koshiyama \\
  Holistic AI \\
  London \\
  UK\\
}

\begin{document}

\maketitle

\begin{abstract}
The use of automated decision tools in recruitment has received an increasing amount of attention. In November 2021, the New York City Council passed a legislation (Local Law 144) that mandates bias audits of Automated Employment Decision Tools. From 15th April 2023, companies that use automated tools for hiring or promoting employees are required to have these systems audited by an independent entity. Auditors are asked to compute bias metrics that compare outcomes for different groups, based on sex/gender and race/ethnicity categories at a minimum. Local Law 144 proposes novel bias metrics for regression tasks (scenarios where the automated system scores candidates with a continuous range of values). A previous version of the legislation proposed a bias metric that compared the mean scores of different groups. The new revised bias metric compares the proportion of candidates in each group that falls above the median. In this paper, we argue that both metrics fail to capture distributional differences over the whole domain, and therefore cannot reliably detect bias. We first introduce two metrics, as possible alternatives to the legislation metrics. We then compare these metrics over a range of theoretical examples, for which the legislation proposed metrics seem to underestimate bias. Finally, we study real data and show that the legislation metrics can similarly fail in a real-world recruitment application. 
\end{abstract}


\section{Introduction}
The fairness of Artificial Intelligence systems has received increased attention in recent years. As more of these systems become ubiquitous in our society, it becomes imperative to ensure that their use is fair towards all members of our society. This is particularly true for sensitive applications that are involved in critical decision-making processes. For example, automated tools are used in healthcare to aid medical diagnosis \cite{McKinney2020}, in finance to predict credit scores \cite{Tsai2008}, and in recruitment to evaluate candidates and predict future job performance \cite{dastin2018amazon}. In these cases, the use of automated tools has the potential of affecting the lives and life chances of the individuals involved. For all of these reasons, there has been an increasing demand for the regulation of AI systems and their applications. For example, the European Commission proposed the Harmonised Rules on Artificial Intelligence, colloquially known as the EU AI Act, which aims to create an ecosystem of trust and regulate the AI tools that are available on the EU market. \\

Taking a more targeted approach, the New York City Council passed Local Law 144 in November 2021, requiring bias audits of automated employment decision tools (AEDTs), thereby narrowing their focus to automation used in talent management. Under the law, Employers  that use AEDTS to screen candidates for employment or employees for promotion in the City must have these systems audited by an independent entity beginning April 15, 2023 (postponed from January 1, 2023). However, one of the key problems in formulating adequate legislation, and in the AI fairness field in general, is to find effective ways to quantify bias. Accordingly, the Department of Consumer and Worker Protection (DCWP) has proposed bias metrics that require bias to be determined using impact ratios that compare outcomes for subgroups based on sex/gender and race/ethnicity categories at a minimum. Whereas the use of disparate impact metrics for binary outputs is quite established in the field of pre-employment testing (\cite{ployhart2008diversity, hough2001determinants, oswald2016measuring}), the legislation proposes novel rules to measure bias for continuous outputs. \\

A first draft of the DCWP's proposed rules required auditors to compare the mean scores of different groups. In contrast, an updated version of the rules proposes that auditors should measure and compare the proportion of candidates in each subgroup that fall above the median for the sample. In this paper, we will present a critical analysis of both metrics and argue that they fail to capture distributional differences over the whole domain, and therefore cannot reliably detect bias. We will show that these metrics underestimate bias over a range of theoretical examples. We suggest two alternative metrics that could be better at capturing the evolution over the whole distribution. Finally, we assess and compare all the metrics on real-world data, obtained from a machine learning system used in recruitment. 

\section{Background}
Artificial intelligence is commonly used in talent management to screen candidates and decide who should advance to the next stage of the recruitment funnel. This type of task is called \textbf{binary classification}, because the aim is to classify the data (candidates, in our example) into unsuccessful (0) and successful (1) classes. Let $X$ denote the random variable associated with the feature vector, and let $f$ be the binary prediction function satisfying $f(X)\in\{0,1\}$ (which may or may not be random).
Also suppose we have a protected attribute $A$ with $K$ possible values encoded as $i=1,\hdots, K$. We assume we have a distribution on $(f(X), A)$. We do not assume we are in possession of a distribution involving the instance $X$ because we usually do not have access to the instances or the model $f$, only its outputs.\\

In binary classification, the most common metric used to measure bias is known as \textbf{disparate impact} \cite{Feldman2014}, which examines whether hiring rates are different across groups. To compute disparate impact, we first compute the selection rate for each group $i$ as $SR_i = \mathbb{P}(f(X)=1|A=i)= \mathbb{E}(f(X)|A=i)$. Note that in binary classification, the selection rate can also be written as an expectation. \\

The Binary Disparate Impact ($DI$) is then computed as the ratio of selection rates between groups, taking the largest as the denominator (so that the computed metric is below 1). 
\begin{equation}
    DI_{i}= \frac{\mathbb{E}(f(X)|A=i)}{\max_j \mathbb{E}(f(X)|A=j)}
\end{equation}

This metric has been extensively used and studied within the computational fairness literature \cite{barocas2017fairness, chouldechova2017fair, feldman2015certifying, zafar2017fairness}. In the case that the selection rates are equal across groups, we can say our classifier satisfies Statistical Parity (or Demographic Parity), which is represented by the following equation:
\[SR_{i}=SR_{j} \ \forall \ i,j \in \{1, \hdots, K\}\]
Statistical Parity is an \textbf{Equality of Outcome} fairness notion, meaning that it seeks to equalize the outcomes of our model for different subgroups of the population \cite{dwork2012fairness, agarwal2018reductions, Agarwal2019}. There are also \textbf{Equality of Opportunity} fairness notions that seek to equalize the performance of our model for different subgroups \cite{hardt2016equality}. In this paper, we focus on Equality of Outcome, because we do not assume we have the true labels that are necessary to compute performance metrics.\\

However, not all applications of Artificial Intelligence in recruitment result in binary outcomes; many systems use \textbf{regression}. In regression, the outputs $f(X)$ of our model are no longer restricted to 0,1 -- they can now take a continuous range of values (e.g., 0 to 100). While it is not immediately clear how the fairness notions from binary classification might generalise to the regression setting, following the definition of \cite{Agarwal2019}, a regressor satisfies Statistical Parity if the outcome $f(X)$ is independent of the protected attribute:
\[\mathbb{P}(f(X)\geq z|A=i)=\mathbb{P}(f(X)\geq z|A=j) , \forall i,j \ \forall z\]
While binary classification allowed us to replace the probability with an expectation, this is not possible in the regression setting. Also note that what was once a single condition now becomes a family of equations that have to be satisfied, indexed by the threshold $z$. This means that the condition for regression data is considerably stronger than the binary one, where fairness is imposed at all possible thresholds. Originally, the metric proposed in the legislation for regression data is taken in analogy to the metric in the binary setting, where impact ratios compared the mean score for different groups. In equations, it is computed as
\begin{equation}
    MeanDI_{i}= \frac{\mathbb{E}(f(X)|A=i)}{\max_j \mathbb{E}(f(X)|A=j)}
\end{equation}
 
This metric was then withdrawn in the updated rules published by the DCWP and an alternative was offered. For a proportion $p$, we denote by $Q(p)$ the corresponding quantile value (function satisfying $P(f(x)\leq Q(p))=p$). The newly proposed metric uses the median value as threshold ($median=Q(0.5)$) to compute a binary disparate impact.
\begin{equation}
    MedDI_{i}=\frac{P(f(X)\geq median|A=i)}{\max_j P(f(X)\geq median|A=j)}
\end{equation}

Note that this metric is equivalent to setting the success rate of the overall sample to $50\%$, and then comparing the success rates of the different subgroups. 

\section{Proposed Metrics}
The metrics we propose in this section are based on reducing the regression case to the binary case. There are different paradigms for thresholding (binarizing) data that we cover in the Discussion section. Here, we will use the paradigm based on quantile values. The reason for this choice is that it reduces the sensitivity of the metric to outliers and makes it so that we can work in the range $[0,1]$ regardless of the original data domain. We first introduce useful notation for the binary disparate impact obtained at a given proportion. For a proportion $p$, we denote by $Q(p)$ the corresponding quantile value (function satisfying $P(f(X)\leq Q(p))=p$).
\begin{equation}
    BinDI_{i}(p)=\frac{P(f(X)\geq Q(p)|A=i)}{\max_j P(f(X)\geq Q(p)|A=j)}
\end{equation}

In this section, we introduce two new metrics: $AucDI$ (area under binary disparate impact curve) and $PfDI$ (probability of fair binary disparate impact). Suppose we start with a prior distribution $\pi(p), p\in[0,1]$ for the proportion of \textbf{unsuccessful} candidates. If we know for certain which proportion $p^*$ will be unsuccessful, we can use a Kronecker delta prior $\pi(p)=\delta_{p^*}(p)$. If we have no prior knowledge on the proportion of rejected candidates, we can use a flat (uniform) prior $\pi(p)=1, p\in[0,1]$. We will only use flat priors throughout the paper for simplicity, but note that using a good prior for the proportion $p$ can drastically improve the quality of the metrics. The equations for these metrics are as follows:
\[AucDI_{i} = \int_{p=0}^1 BinDI_{i}(p) \pi(p) dp\]
and
\[PfDI_{i} = \int_{p=0}^1 \mathbbm{1}[BinDI_{i}(p)\geq0.8] \pi(p) dp\]
In the above equation, $\mathbbm{1}[\textrm{condition}]$ denotes the indicator function on the given condition (function that is 1 if condition holds, 0 if not). The above metrics offer two different ways of aggregating the data obtained from the binarization procedure into a metric for regression data. Note that we used disparate impact based metrics, to match the conventions of the US legislation. Nothing prevents us from aggregating other binary classification bias metrics (e.g., statistical parity difference) in the same ways. In the next sections, we compare the metrics proposed by the DCWP with the metrics proposed in this paper and look for differences in their properties.
\section{Fairness Bounds}
Local Law 144 does not explicitly indicate a threshold for fairness, neither for binary classification nor regression metrics. It only mandates that the metrics be computed for each protected attribute, as well as intersectional subgroups. In recruitment, when it comes to binary classification, it is common to refer to the four-fifth rule, which indicates that the impact ratio should be larger than 4/5 (0.8) for the selection rates to be considered fair (US Equal Employment Opportunity Commission, EEOC, 1979). While the bound isn't stated to be the same for regression disparate impact metrics, one can extrapolate that we would use the same bound. \\ 

What about bounds for $AucDI$ and $PfDI$? A perfectly fair regressor satisfying Statistical Parity will have both $AucDI$ and $PfDI$ of 1, whereas the minimum value for these metrics is 0. Even though there is no perfect way of choosing a fairness threshold, we will consider here a fairness threshold of 0.8 for both metrics. In the case of $AucDI$, this means the average Binary Disparate Impact over the whole range of possible proportions should be no less than 0.8. In the case of $PfDI$, this means that by taking a random proportion in $[0,1]$, the probability of the binary data produced being fair is above 0.8. However, it is worth noting that alternative thresholds are possible. 

\section{Experiments}
\subsection{Theoretical Examples}
In this section, we give some examples to motivate the argument that $MeanDI$ and $MedDI$ are not reliable measurements of bias. For the ratio of averages ($MeanDI$), the reason lies in the way this metric uses the average as a descriptor of a distribution. By reducing distributions to their average value, we throw away a lot of valuable distributional information. For the median-based thresholding ($MedDI$), the reason is that considering only one threshold is not enough to get a full distributional picture. Distributions are infinite dimensional objects, so they are difficult to describe with a scalar value. Of course, all metrics involve reduction from a high information object to a low information object (usually to a scalar). That is why metrics must be carefully designed to measure a specific notion. Even then, a single notion will usually have different metrics that measure it, each of which uses a slightly different lens. \\

To get a better idea of the way bias changes across the whole  distribution, we use  binary disparate impact plots. In a binary disparate impact plot, we threshold the regression scores at different quantile values, and then plot the binary disparate impact as a function of the various proportion quantiles. Essentially, we can look at how the binary disparate impact varies when we select all the candidates, $90\%$ of the candidates, $80\%$ of the candidates and so forth until none of the candidates are selected. If the regression scores for the two groups considered have exactly the same distribution, then we get perfect fairness at every threshold (a flat disparate impact curve constant at 1). In practice, however, the distributions will often differ, and the binary disparate impact curve will not be fully flat. Usually, the curve will vary in and out of the fairness range. The only properties that all binary disparate impact curves share is that they have a similar behaviour at the extremes, that is, they always contain the points $(0,1)$ and $(1,0)$ because at quantile 0 all candidates succeed and at quantile 1 none of the candidates succeed. \\

Notably, the middle point of the binary disparate impact plot corresponds to selecting half of the candidates and is thus equivalent to $MedDI$, the metric proposed for Local Law 144. The binary disparate impact plot provides us with an outlook over the whole distribution, rather than just one point. We will see that in many cases, even if $MedDI$ yields a fair result, the binary disparate impact curve might speak a different story. We will also argue that our metrics, $AucDI$ and $PfDI$, are better suited to capture bias over the whole distribution. \\

\subsubsection{Example 1}

\begin{figure}
\begin{subfigure}{\linewidth}
\centering
\includegraphics[width=10cm]{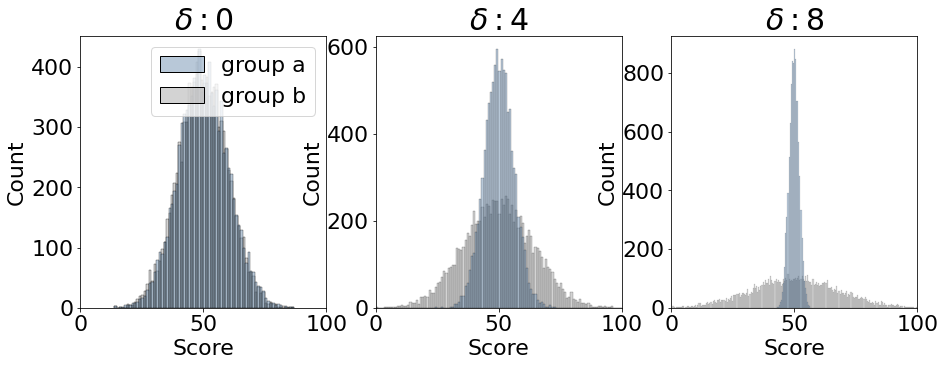}
\caption{Distributions}
\label{fig:1sub1}
\end{subfigure}\\[1ex]
\begin{subfigure}{\linewidth}
\centering
\includegraphics[width=10cm]{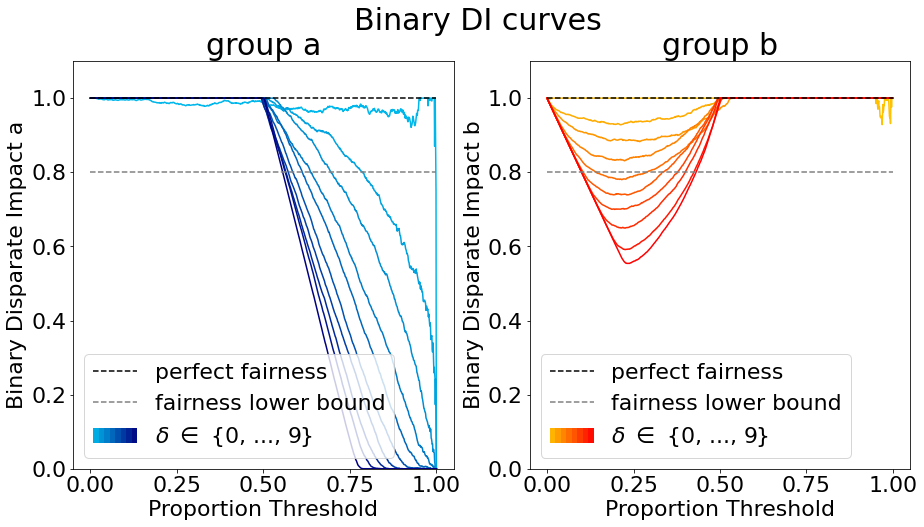}
\caption{Binary Disparate Impact Curves}
\label{fig:1sub2}
\end{subfigure}\\[1ex]
\begin{subfigure}{\linewidth}
\centering
\includegraphics[width=10cm]{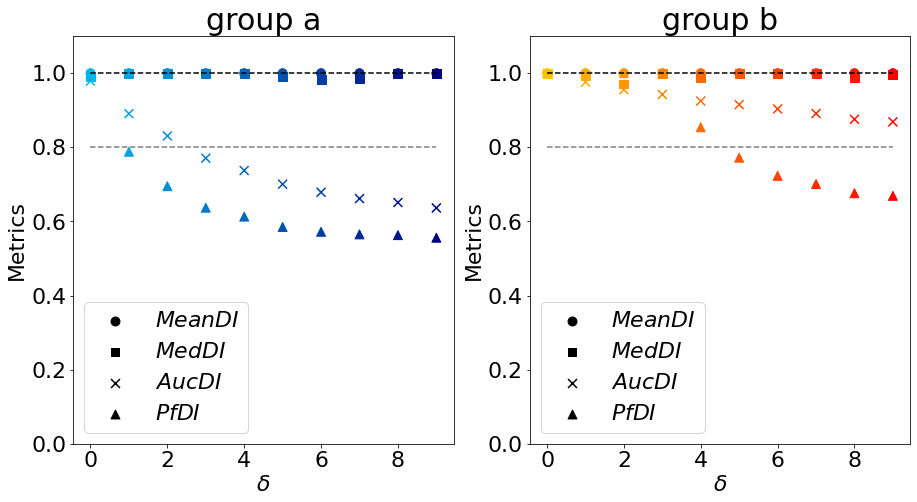}
\caption{Metrics}
\label{fig:1sub3}
\end{subfigure}
\caption{\textbf{Example 1}. The first example consists of a $N(50,10-\delta)$ distribution for group $a$ and a $N(50,10+\delta)$ for group $b$, where we let the parameter $\delta$ have range $\{0,\hdots,9\}$. \textbf{(a) Distributions}. We show the distributions of group $a$ and group $b$ for parameter $\delta \in \{0, 4, 8\}$. Observe how the distribution of group $a$ narrows while that of group $b$ spreads with increasing $\delta$. \textbf{(b) Binary Disparate Impact Curves}. We show the evolution of the binary disparate impact as we change the proportion threshold from 0 to 1, for groups $a$ and $b$. \textbf{(c) Metrics}. We plot the metrics $MeanDI, MedDI, AucDI, PfDI$ as a function of parameter $\delta$, for groups $a$ and $b$.}
\label{fig:test1}
\end{figure}

As a first example, take the distributions of groups $a$ and $b$ to be normal with mean $\mu=50$. These distributions only differ in their standard deviations $\sigma_a$ and $\sigma_b$. We create a parameter range using an extra parameter $\delta$ to regulate the standard deviation difference. We set $\sigma_a=10-\delta$, and $\sigma_b=10+\delta$. The distribution of group $a$ gets increasingly narrow around 50 as $\delta$ increases while the distribution of group $b$ spreads out (Fig. \ref{fig:1sub1}). Clearly, for $\delta=0$ the distributions will be identical and thus we expect to find perfect fairness. As $\delta$ increases, the distributions diverge and the fairness properties change. For large values of $\delta$, we will have candidates from group $a$ score consistently around 50, while group $b$ scores will be more evenly distributed over the whole range. Whether this is fair or not will strongly depend on the choice of binarization threshold. If we choose a small threshold, then more of the selected candidates will belong to group $a$, and the binary disparate impact will detect bias against group $b$ (Fig. \ref{fig:1sub2}). On the other hand, higher thresholds result in bias against group $a$ (Fig. \ref{fig:1sub2}).
Since the two groups have the same median and mean, even for varying parameter $\delta$, the value of the metrics $MeanDI$ and $MedDI$ will be $1$ for any $\delta$. $MeanDI$ compares the means of the two groups and so it cannot capture the change in standard deviation, while $MedDI$ looks only at one data point on the binary disparate impact curve. On the other hand, $AucDI$ and $PfDI$ find increasing bias as $\delta$ increases (Fig. \ref{fig:1sub3}). This is because these metrics take into account the whole distribution of scores, and thus are able to capture the dependency of the fairness properties on the  parameter $\delta$. Furthermore, $PfDI$ flags unfair outcomes for both group $a$ and group $b$, for large values of $\delta$. This is because for lower thresholds group $a$ is advantaged, whereas larger thresholds advantage group $b$.

\subsubsection{Example 2} 
\begin{figure}
\begin{subfigure}{\linewidth}
\centering
\includegraphics[width=10cm]{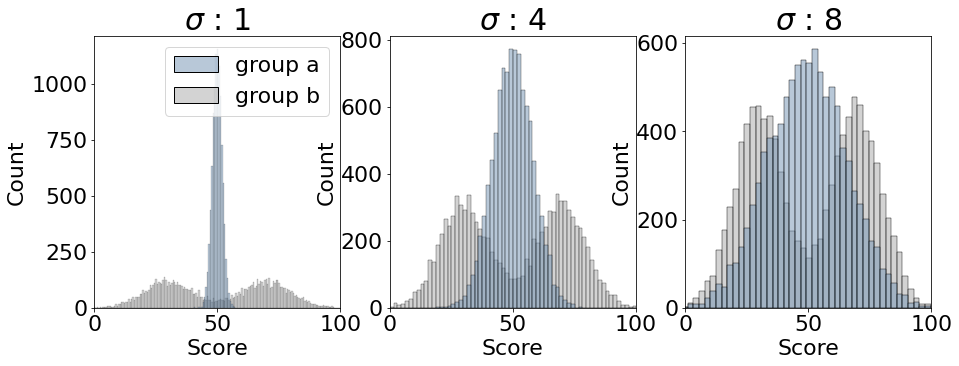}
\caption{Distributions}
\label{fig:2sub1}
\end{subfigure}\\[1ex]
\begin{subfigure}{\linewidth}
\centering
\includegraphics[width=10cm]{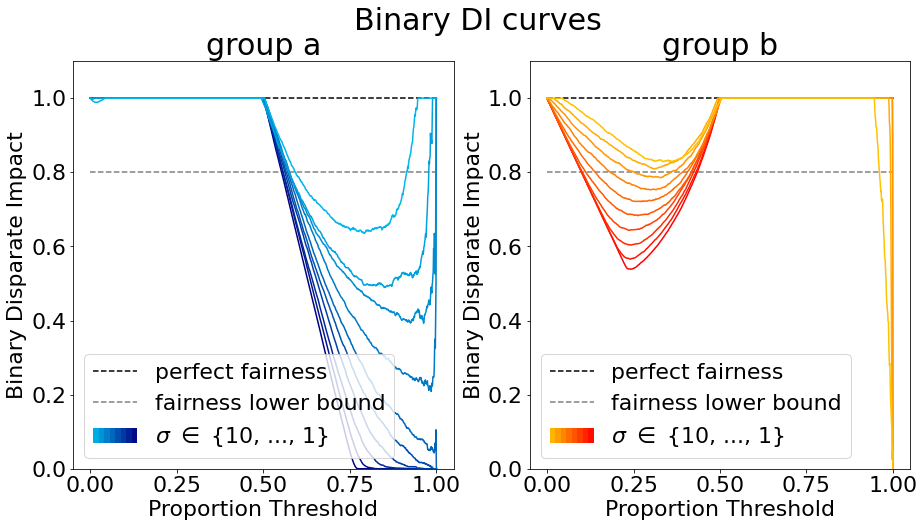}
\caption{Binary Disparate Impact Curves}
\label{fig:2sub2}
\end{subfigure}\\[1ex]
\begin{subfigure}{\linewidth}
\centering
\includegraphics[width=10cm]{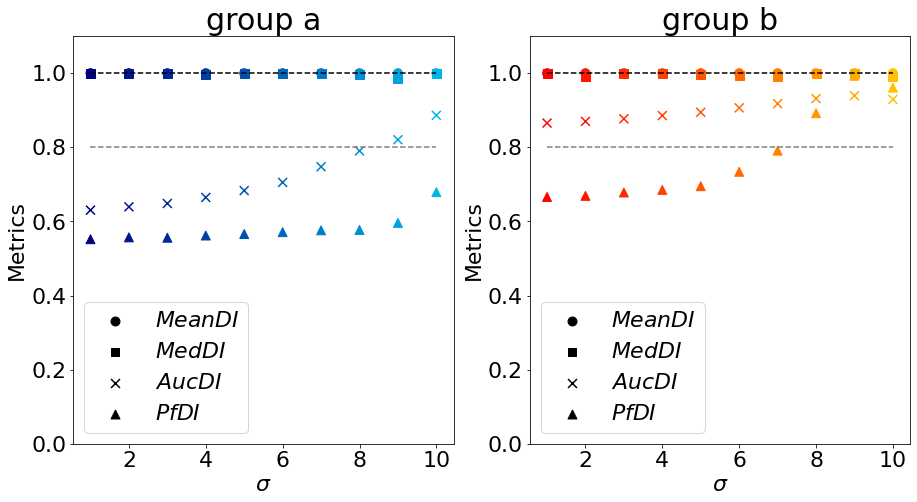}
\caption{Metrics}
\label{fig:2sub3}
\end{subfigure}
\caption{\textbf{Example 2}. The second example consists of a $N(50,\sigma)$ distribution for group $a$ and a bimodal mixture of $N(20,10)$ and $N(80,10)$ for group $b$. We let the parameter $\sigma$ have range $\{10,\hdots,1\}$. \textbf{(a) Distributions}. We show the distributions of group $a$ and group $b$ for parameter $\sigma \in \{1, 4, 8\}$. Observe how the distribution of group $a$ narrows while that of group $b$ stays fixed with decreasing $\sigma$. \textbf{(b) Binary Disparate Impact Curves}. We show the evolution of the binary disparate impact as we change the proportion threshold from 0 to 1, for groups $a$ and $b$. \textbf{(c) Metrics}. We plot the metrics $MeanDI, MedDI, AucDI, PfDI$ as a function of parameter $\delta$, for groups $a$ and $b$.}
\label{fig:test2}
\end{figure}

As a second example, we take one normal distribution for group $a$ centered at 50 with standard deviation $\sigma$, and a bimodal mixture of two normals centered at 20 and 80 for group $b$ with standard deviation equal to 10 in both cases (see Fig. \ref{fig:2sub1}). 
This example is notable for three reasons. Firstly, candidates from group $b$ score at either the lower or higher end of the value range, while group $a$ scores consistently around the middle point. This means that for a sufficiently small threshold, only candidates from group $b$ will be rejected. Conversely, for a sufficiently large threshold, only candidates from group $b$ will be selected. Fairness is  therefore highly dependent on the binarization threshold: lower thresholds result in bias against group $b$, while higher thresholds result in bias against group $a$ (Fig. \ref{fig:2sub2}).
Secondly, fairness depends on the standard deviation parameter $\sigma$. For smaller values of $\sigma$, the disparity between groups is more pronounced. When $\sigma$ increases, the overlap between the two distributions increases, resulting in a fairer outcome (Fig. \ref{fig:2sub2}). 
Thirdly, we can see that group $a$ and group $b$ have the same  mean and median, independently of the choice of the parameter $\sigma$. The regression metrics proposed by the legislation, $MeanDI$ and $MedDI$, will therefore be consistently equal to 1 (Fig. \ref{fig:2sub3}). $MeanDI$ fails at detecting any bias because the group mean cannot capture the spread of the distribution. $MedDI$ is lacking in that it only considers the middle point on the binary disparate impact curve, which happens to be fair in this example. On the other hand, $AucDI$ and $PfDI$ are able to capture the trend over the whole distribution. Under our settings, $AucDI$ and $PfDI$ detect bias mostly against group $a$, and both metrics yield fairer values as the standard deviation $\sigma$ increases. Interestingly, for lower values of $\sigma$, $PfDI$ falls under the fairness threshold for both group $a$ and group $b$. $PfDI$ is, therefore, able to detect bias against both groups at the same time.

\subsubsection{Example 3} 

\begin{figure}
\begin{subfigure}{\linewidth}
\centering
\includegraphics[width=10cm]{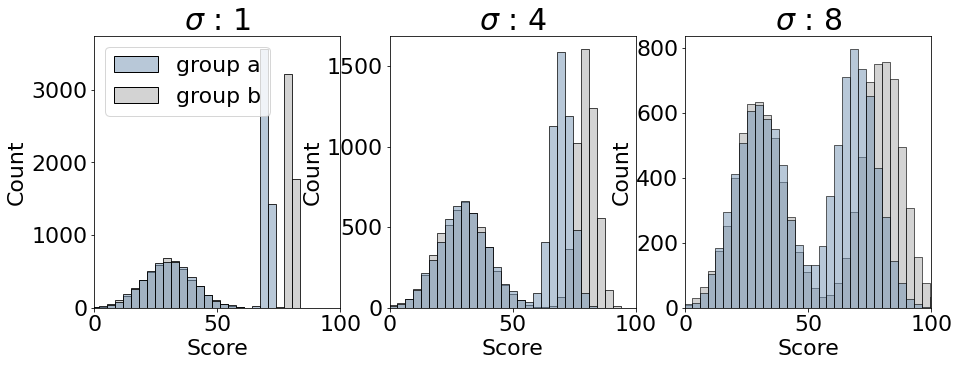}
\caption{Distributions}
\label{fig:3sub1}
\end{subfigure}\\[1ex]
\begin{subfigure}{\linewidth}
\centering
\includegraphics[width=10cm]{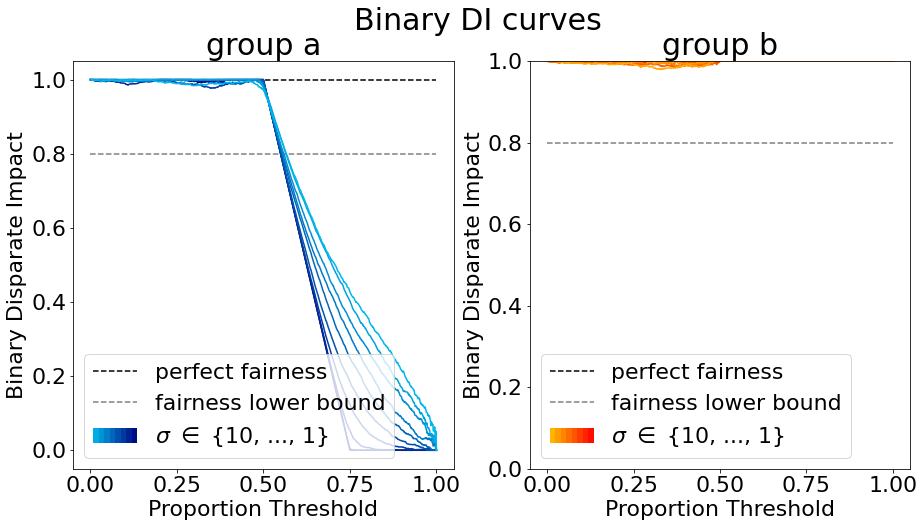}
\caption{Binary Disparate Impact Curves}
\label{fig:3sub2}
\end{subfigure}\\[1ex]
\begin{subfigure}{\linewidth}
\centering
\includegraphics[width=10cm]{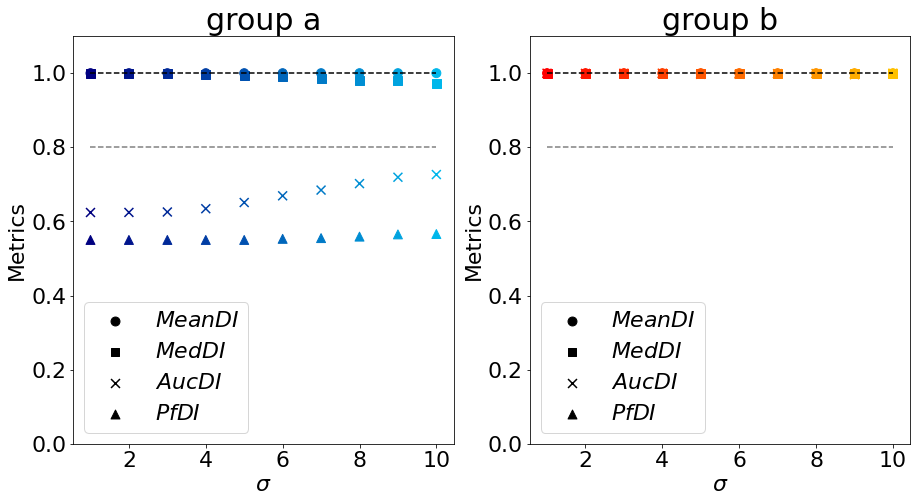}
\caption{Metrics}
\label{fig:3sub3}
\end{subfigure}
\caption{\textbf{Example 3}. The third example consists of a bimodal mixture of $N(30,10)$ and $N(70,\sigma)$ for group $a$ and a bimodal mixture of $N(30,10)$ and $N(80,\sigma)$ for group $b$. We let the parameter $\sigma$ have range $\{10,\hdots,1\}$. \textbf{(a) Distributions}. We show the distributions of group $a$ and group $b$ for parameter $\sigma \in \{1, 4, 8\}$. \textbf{(b) Binary Disparate Impact Curves}. We show the evolution of the binary disparate impact as we change the proportion threshold from 0 to 1, for groups $a$ and $b$. \textbf{(c) Metrics}. We plot the metrics $MeanDI, MedDI, AucDI, PfDI$ as a function of parameter $\delta$, for groups $a$ and $b$.}
\label{fig:test3}
\end{figure}

As a third example, we take bimodal distributions for both groups $a$ and $b$. The bimodal distribution for group $a$ has modes at 30 and 70 with standard deviations 10 and $\sigma$, the bimodal distribution for group $b$ has modes at 30 and 80 with standard deviations 10 and $\sigma$ (Fig. \ref{fig:3sub1}). In this example, the lower half of each group have the same distribution. However, the upper halves of the two groups are centered at different values (70 for group $a$, 80 for group $b$). This means that candidates in the top half of the distribution score quite differently depending on the group they belong to, and this difference gets larger for lower values of $\sigma$. The overall data is set up so as to have mean and median values close to 50 and all distributional bias occurs after the value 50. Hence, the metrics $MeanDI$ and $MedDI$, cannot detect bias and are close to 1 for any value of the parameter $\sigma$.  
On the other hand, we begin to see distributional bias against group $a$ as soon as our threshold exceeds 50 (Fig. \ref{fig:3sub2}). This shows as a sharp decrease in the binary disparate impact curve after proportion 0.5. We can see that fairness is strictly related to the choice of binarization. Since $AucDI$ and $PfDI$ take into account the whole distributional range, they consistently flag these pairs of distributions as biased against group $a$, for any $\sigma$, with bias increasing as $\sigma$ decreases  (Fig.\ref{fig:3sub3}).

\subsection{Real Data}

The previous examples served as a proof of principle; they showed that it is important that regression metrics consider the trend over the whole output distribution, rather than only one data point. We will next show how similar issues can present in real-world data as well. We analysed data from a recruitment company that used machine learning systems to score candidates on a scale from 0 to 100. This was an interesting example for us, since the company ended up binarizing these scores by thresholding at score 50 (in the middle of the 0 to 100 range).\\

They trained 10 different models, which we named predictor0, predictor1, ..., predictor9. In addition to the scores, the company provided us with ethnicity and gender information for most of the candidates (77460 after dropping NaN values). The gender attribute contains the Male and Female subgroups, while the ethnicity contains the Asian, Black,  Hispanic/Latino, White and Two or more ethnicities (Two+) subgroups. We name the disparate impact obtained by binarizing at 50, $ThreshDI$, to respect our naming convention. We decided to analyse this case in order to compare the bias metrics calculated on the regression scores to the final actual binarization used. \\

\begin{figure}
     \centering
     \begin{subfigure}[b]{0.45\textwidth}
         \centering
         \includegraphics[width=6.5cm]{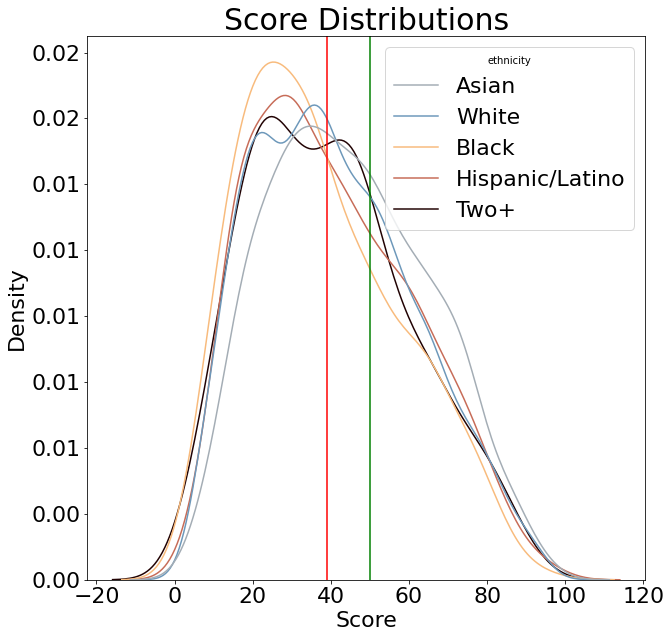}
         \caption{Data distributions}
         \label{fig:real1sub1}
     \end{subfigure}
     \begin{subfigure}[b]{0.45\textwidth}
         \centering
         \includegraphics[width=6.2cm]{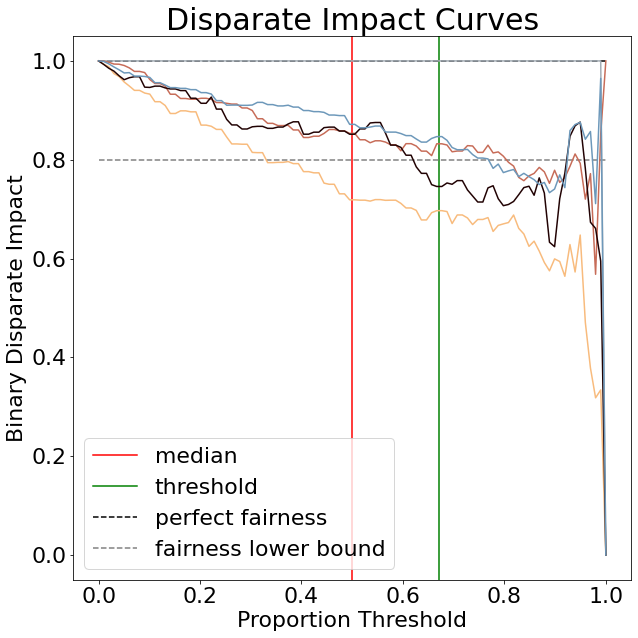}
         \caption{Binary Disparate Impact curves}
         \label{fig:real1sub2}
     \end{subfigure}
     \hfill
    \caption{\textbf{Real Data Example}. This example consists of real company regression data (predictor3), ranging from 0 to 100. We study this example with the ethnicity protected attribute. \textbf{(a) Data Distribution}. We show the (kde estimated) distributions of all groups (Asian, White, Black, Hispanic/Latino, Two+). \textbf{(b) Binary Disparate Impact curves}. We plot the binary disparate impact for each group as a function of the proportion threshold.}
    \label{fig:real1}
\end{figure}

Given that we have 10 predictors and 2 protected attributes, there are 20 settings in total. We calculated both the binary disparate impact obtained at the threshold of 50 ($ThreshDI$) and that obtained at the median ($MedDI$). We considered the metrics to detect bias if they fell under the 0.8 fairness threshold. We scanned the settings looking for cases where $ThreshDI$ and $MedDI$ disagree (one violates the four-fifths rule and the other does not) for at least one of the relevant subgroups. Note that we are comparing a binarization at the median and at score 50 (which is right in the middle of the range). If the data distribution was symmetric and centered in the middle range (like a Gaussian, for example), then these two metrics would yield the same result. We, therefore, did not expect the metrics to differ substantially. Nonetheless, we found 4 
cases where bias was masked, with  $ThreshDI$ detecting bias and $MedDI$ not for 3 of these cases. Interestingly, $MeanDI$ undervalued bias on all of these cases (Tables in Appendix \ref{appendix:tables}). One of these examples, predictor3 with the ethnicity protected attribute, is displayed in Fig. \ref{fig:real1}. As observed in the metrics Table  \ref{tab:realmetrics1}, the subgroups Black and Two+ are flagged as biased by some metrics. The Black group is flagged by $MedDI$, $ThreshDI$, $AucDI$ and $PfDI$, while the Two+ group is flagged by $ThreshDI$ and $PfDI$. We can see from the Disparate Impact curves (Fig. \ref{fig:real1}) that binarizing at the median (red line) or threshold (green line) fall respectively in and out of the fairness threshold for the subgroup Two+.
 \\

\begin{table}[h]
\centering
\begin{tabular}{lccccc}
\toprule
{} &  Asian &     Black &  Hispanic/Latino &      Two+ &     White \\
\midrule
$MeanDI$ &   1.00 &  0.840863 &         0.908923 &  0.892794 &  0.918503 \\
$MedDI$  &   1.00 &  \textcolor{red}{0.718925} &         0.853346 &  0.851637 &  0.871630 \\
$ThreshDI$  &   1.00 &  \textcolor{red}{0.696958} &         0.832264 &  \textcolor{red}{0.745903} &  0.846828 \\
$AucDI$  &   0.99 &  \textcolor{red}{0.744136} &         0.860472 &  0.828036 &  0.866973 \\
$PfDI$   &   0.99 &  \textcolor{red}{0.330000} &         0.830000 &  \textcolor{red}{0.650000} &  0.830000 \\

\bottomrule
\end{tabular}
\caption{Metrics Table. We display all computed metrics for each subgroup of the ethnicity attribute, for the scores of predictor3. Metrics with values below 0.8 (flag) and we denote that by marking their value in red.}
\label{tab:realmetrics1}
\end{table}

We should note here that we are not claiming that our bias metrics can always detect bias. Indeed, the concepts of bias and fairness are multi-faceted and complex. The fairness of a regression model will depend on how the score is taken into account in the final decision. For example, if we know that a specific threshold will be used, like in the data analysed, it would be preferable to treat the problem as a binary classification task. However, in cases where there is no final thresholding or the threshold is undefined, we argue that it is indeed preferable to use metrics that take into account the whole distribution. The metrics we proposed are just instances of many possible metrics, but we suggest that they are better indicators of the fairness of the data as a whole (as opposed to the legislation metrics). We list the observations obtained from our experiments below:
\begin{itemize}
    \item The DCWP's original regression disparate impact $MeanDI$ based on the ratio of means seems to consistently under-detect bias.
    \item $MedDI$ is not sufficient by itself to get a picture of the fairness properties of the data. We find real-world cases where bias is revealed only at certain thresholds.
    \item $PfDI$ seems to flag more often than other metrics and take on more extreme values, both for fair and unfair ranges. 
    \item $AucDI$ seems to be able to detect distributional bias. Adding an appropriate prior will make the metric even more meaningful.
    \item It would be interesting to study the behaviour of both $PfDI$ and $AucDI$ further and deduce the most meaningful choice of threshold for each one. 
    \item In our opinion, computing multiple metrics is the best option to get a fuller picture of the fairness properties of the data.
\end{itemize}

\section{Discussion}

In this paper, we have argued that the proposed metrics for determining whether regression systems are biased under NYC Local Law 144 are not reliable in detecting bias. Since $MeanDI$ uses the mean as a descriptor of each distribution, this metric will usually disguise the intricacies of distributional information, and consistently underestimate bias. Indeed, we have shown that this metric almost never triggers, even in cases that are clearly biased. The $MedDI$ metric offers an improvement upon the original proposal and is equivalent to fixing the success rate of the whole population to $50\%$ and then comparing the success rates of each group. Yet, we argue that looking at one threshold ($50\%$) is not enough to capture distributional bias. For instance, bias may only reveal itself at higher thresholds. To solve this issue, we propose two alternative metrics that function by aggregating bias information at every possible threshold: $AucDI$ and $PfDI$. $AucDI$ functions by aggregating the information over all proportion thresholds with an area under the curve approach, while $PfDI$ can be thought of as a probability of obtaining fair binary data ($DI\geq0.8$) for a randomly chosen threshold. As mentioned above, we include the option to add a prior in both proposed metrics, encoding any information we have about the proportion of rejected/accepted candidates. It would also be interesting to explore different priors and fairness bounds to see how it would change the results for these metrics.\\

It is important to note that we chose to look at quantiled domains in both of these metrics (domain that encodes proportion of rejected applicants). The reason behind the choice is that a quantiled range makes the metrics more robust to outliers. We discuss two alternatives to this choice. We could, for example, binarize based on score, similar to the $ThreshDI$ metric from the real data example. In this case, we would choose some threshold $z$ from the original data domain, and consider all scores above $z$ successful and all scores below $z$ unsuccessful. Otherwise, we could binarize by selecting the top $K$ candidates and set everyone else's labels to 0. Both of these alternative paradigms would lead to differences in the way we define and compute the metrics. Which paradigm is most fit to describe a situation will crucially depend on context, and there is no a-priori way to decide which to use.\\

Furthermore, all of the metrics discussed thus far are based on thresholding the data. However, there are two other types of metrics that can be suited for regression data and are worth discussing. The first is \textbf{ranking}-based bias metrics. These are useful in the setting where the scores are used to create a ranking of the candidates. The paradigm in ranking is that the utility (benefit) will go down as a function of rank, so that ranking first is better than second and so forth. We can think of candidates coming up on a webpage in order. In this case, being candidate number 100 means the chances of being selected are almost null. A number of such metrics have been introduced and discussed in the literature \cite{rankingbias, Geyik2019, beutel2019fairness, raj2020comparing}. Finally, we would like to mention that another approach to assessing the fairness of regression data can be found in \textbf{statistical testing}. For example, the  Kolmogorov-Smirnov (KS) test \cite{kolmogorov1933sulla} can be used to compare $F_a$ and $F_b$, the empirical cumulative distributions for the two groups, by computing the supremum of their difference $D = \sup_{x}|F_a(x)-F_b(x)|$. If this difference is large enough for the null hypothesis to be rejected, the model can be assumed to be unfair. If the difference is not large enough, the test is inconclusive. In practice, this test will require a large enough sample size to achieve appropriate confidence levels, but it is a good custom to use a statistical test, even after a dataset has been flagged as biased by a bias metric \cite{taskesen2021statistical}. \\

\bibliographystyle{unsrt}  
\bibliography{bib}  

\appendix

\section{Real Data Experiments (further Plots and Metrics Tables)}
\label{appendix:tables}
\begin{figure}[b]
         \centering
         \includegraphics[width=10cm]{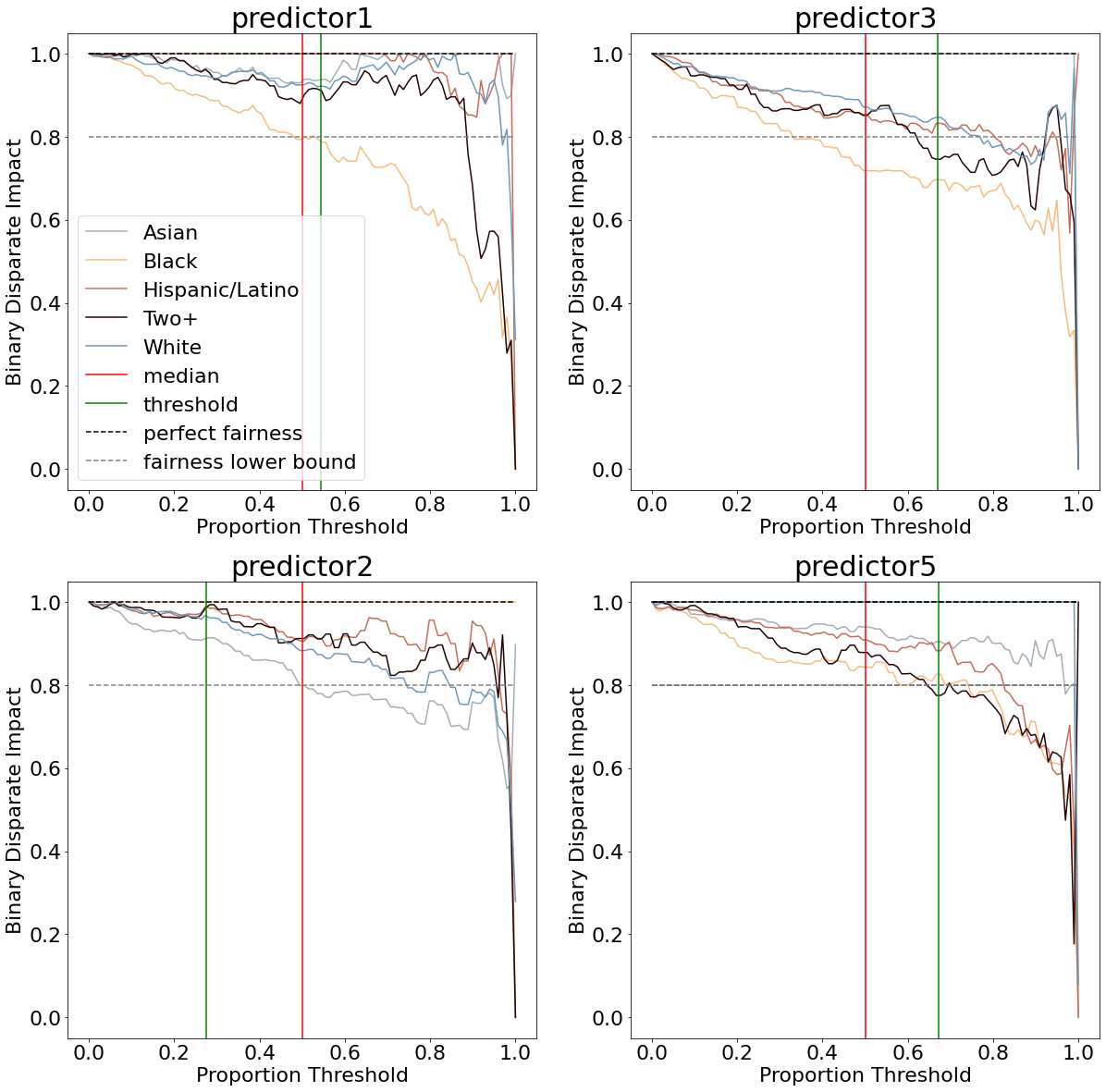}
         \caption{\textbf{Binary disparate impact curves.} We show the binary disparate impact curves for all 4 examples where $MedDI$ and $ThreshDI$ disagreed (one flags and other doesn't) for at least one group.}
         \label{fig:allcases}
\end{figure}

\begin{table}[h]
\centering
\begin{tabular}{lccccc}
\toprule
{} &     Asian &     Black &  Hispanic/Latino &      Two+ &     White \\
\midrule
$MeanDI$ &  0.980857 &  0.861800 &          1.00000 &  0.951457 &  0.966043 \\
$MedDI$  &  0.938114 &  0.800478 &          1.00000 &  0.901684 &  0.925310 \\
$ThreshDI$  &  0.938070 &  \textcolor{red}{0.786071} &          1.00000 &  0.910472 &  0.922407 \\
$AucDI$  &  0.975231 &  \textcolor{red}{0.760605} &          0.97396 &  0.886844 &  0.943328 \\
$PfDI$   &  1.000000 &  \textcolor{red}{0.500000} &          0.99000 &  0.880000 &  0.970000 \\

\bottomrule
\end{tabular}
\caption{Metrics Table (predictor 1 / ethnicity)}
\label{tab:realmetrics1e}
\end{table}

\begin{table}[h]
\centering
\begin{tabular}{lccccc}
\toprule
{} &     Asian &     Black &  Hispanic/Latino &      Two+ &     White \\
\midrule
$MeanDI$ &  0.931422 &  1.000000 &         0.977229 &  0.969312 &  0.962149 \\
$MedDI$  &  \textcolor{red}{0.798804} &  1.000000 &         0.906356 &  0.912210 &  0.883028 \\
$ThreshDI$  &  0.911784 &  1.000000 &         0.988763 &  0.984916 &  0.967474 \\
$AucDI$  &  0.828851 &  0.999598 &         0.924852 &  0.905651 &  0.878264 \\
$PfDI$   &  \textcolor{red}{0.500000} &  1.000000 &         0.960000 &  0.960000 &  \textcolor{red}{0.790000} \\

\bottomrule
\end{tabular}
\caption{Metrics Table (predictor 2 / ethnicity)}
\label{tab:realmetrics2e}
\end{table}

\begin{table}[h]
\centering
\begin{tabular}{lccccc}
\toprule
{} &  Asian &     Black &  Hispanic/Latino &      Two+ &     White \\
\midrule
$MeanDI$ &   1.00 &  0.840863 &         0.908923 &  0.892794 &  0.918503 \\
$MedDI$  &   1.00 &  \textcolor{red}{0.718925} &         0.853346 &  0.851637 &  0.871630 \\
$ThreshDI$  &   1.00 &  \textcolor{red}{0.696958} &         0.832264 &  \textcolor{red}{0.745903} &  0.846828 \\
$AucDI$  &   0.99 &  \textcolor{red}{0.744136} &         0.860472 &  0.828036 &  0.866973 \\
$PfDI$   &   0.99 &  \textcolor{red}{0.330000} &         0.830000 &  \textcolor{red}{0.650000} &  0.830000 \\

\bottomrule
\end{tabular}
\caption{Metrics Table (predictor 3 / ethnicity)}
\label{tab:realmetrics3e}
\end{table}

\begin{table}[h]
\centering
\begin{tabular}{lccccc}
\toprule
{} &  Asian &     Black &  Hispanic/Latino &      Two+ &     White \\
\midrule
$MeanDI$ &  0.949931 &  0.884560 &         0.928716 &  0.902873 &  1.000000 \\
$MedDI$  &  0.939701 &  0.843386 &         0.908458 &  0.878517 &  1.000000 \\
$ThreshDI$  &  0.904694 &  0.825959 &         0.882696 &  \textcolor{red}{0.775178} &  1.000000 \\
$AucDI$  &  0.916716 &  0.816245 &         0.869071 &  0.838914 &  0.990665 \\
$PfDI$   &  0.970000 &  \textcolor{red}{0.720000} &         0.820000 &  \textcolor{red}{0.660000} &  0.990000 \\

\bottomrule
\end{tabular}
\caption{Metrics Table (predictor 5 / ethnicity)}
\label{tab:realmetrics5e}
\end{table}

\end{document}